\documentclass[conference]{IEEEtran}
\IEEEoverridecommandlockouts
\usepackage{cite}
\usepackage{amsmath,amssymb,amsfonts}
\usepackage{algorithmic}
\usepackage{graphicx}
\usepackage{textcomp}
\usepackage{xcolor}
\usepackage{balance}
\def\BibTeX{{\rm B\kern-.05em{\sc i\kern-.025em b}\kern-.08em
    T\kern-.1667em\lower.7ex\hbox{E}\kern-.125emX}}
\begin{document}

\title{Decoding Imagined Speech: A Strictly Subject-Independent Approach Using EEG\\

}

\author{\IEEEauthorblockN{1\textsuperscript{st} Frederik Møllskov Trier}
\IEEEauthorblockA{\textit{Department of Health Technology} \\
\textit{Technical University of Denmark}\\
Ørsteds Plads, 2800  \\
Kgs. Lyngby, Denmark \\
s211923@student.dtu.dk}
\and
\IEEEauthorblockN{2\textsuperscript{nd} Xiaopeng Mao}
\IEEEauthorblockA{\textit{Department of Health Technology} \\
\textit{Technical University of Denmark}\\
Ørsteds Plads, 2800  \\
Kgs. Lyngby, Denmark \\
xiama@dtu.dk}
\and
\IEEEauthorblockN{3\textsuperscript{rd} Sadasivan Puthusserypady}
\IEEEauthorblockA{\textit{Department of Health Technology} \\
\textit{Technical University of Denmark}\\
Ørsteds Plads, 2800  \\
Kgs. Lyngby, Denmark \\
sapu@dtu.dk}
}

\maketitle

\begin{abstract}
Imagined speech decoding from electroencephalography (EEG) has gained increasing attention as a potential communication pathway for individuals with severe motor impairments, yet reported performance often relies on evaluation protocols that do not clearly reflect cross-subject generalization. This study presents a transparent baseline investigation of a multi-class imagined speech EEG dataset under a strictly subject-independent evaluation framework. Two preprocessing and feature extraction pipelines were compared: a time-domain statistical feature approach and a frequency-domain spectral band-power approach, evaluated using subject-wise cross-validation and trial-level majority voting with a random forest classifier. The spectral pipeline achieved a significantly higher mean trial-wise accuracy than the statistical pipeline (49.03 ± 4.18\% vs. 37.97 ± 3.79\%) for coarse-level classification across subjects. Forward feature selection further indicated that a limited subset of frequency bands captured most of the discriminative information. Overall, this work provides a strong basis for future brain-computer interface studies targeting improved cross-subject generalization in EEG-based imagined speech decoding.
\end{abstract}


\section{Introduction}

Communication is central to human life, and the inability to express one-self can be extremely challenging. Patients suffering from locked-in syndrome have functioning cognition, but very limited motor control, resulting in severely restricted communication  \cite{Locked}. Imagined speech decoding has been proposed as a potential future solution, with the idea revolving around translation of brain activity into text or symbols as a means of communication. However, this technology is far from being implemented in real life situations due to the complexity of brain signals and the inherent low signal-to-noise ratio in electroencephalography (EEG) \cite{Lopez}. Decoding brain activity is a complex task that relies on the quality of the input features to a given machine learning (ML) model \cite{Review}.

The publicly available imagined speech EEG dataset introduced by Kumar et al.\cite{Kumar}, commonly referred to as Kumar’s EEG dataset, has become a widely used benchmark. The dataset contains three "coarse-grained" classes (letters, digits, and objects), each further divided into ten "fine-grained" subclasses. In the original study, simple statistical time-domain features extracted from minimally preprocessed EEG were classified using a random forest (RF) model, achieving an accuracy of 85.20\% and 67.03\% at the coarse and fine-grained classification levels, respectively. Subsequent studies have applied more advanced models, including convolutional neural networks \cite{Tirupattur,Tripathi} and transformer-based architectures \cite{Gallo}, both of which outperform the original approach, with the latter achieving the highest reported fine-grained accuracy of approximately 97\%.  However, recent work on EEG-to-text decoding has shown that reported performance can be substantially inflated by leaky evaluation schemes, leading to unreliable assessments of model capability \cite{Jo2024EEGtoText}. Despite its widespread adoption, the Kumar dataset is often evaluated under validation protocols that are not described in sufficient detail, which limits comparability across studies. In particular, it is unclear how the existing works have conducted the data split, i.e., whether the data are split at the window or trial level and in a subject-dependent or subject-independent manner. As a result, the generalizability of the existing works remains uncertain despite their great performances. For practical brain–computer interface (BCI) applications, generalization across subjects is particularly important and appropriately assessed using strictly subject-independent evaluation schemes.

\begin{figure*}[t]
    \centering
    \includegraphics[width=0.95\textwidth]{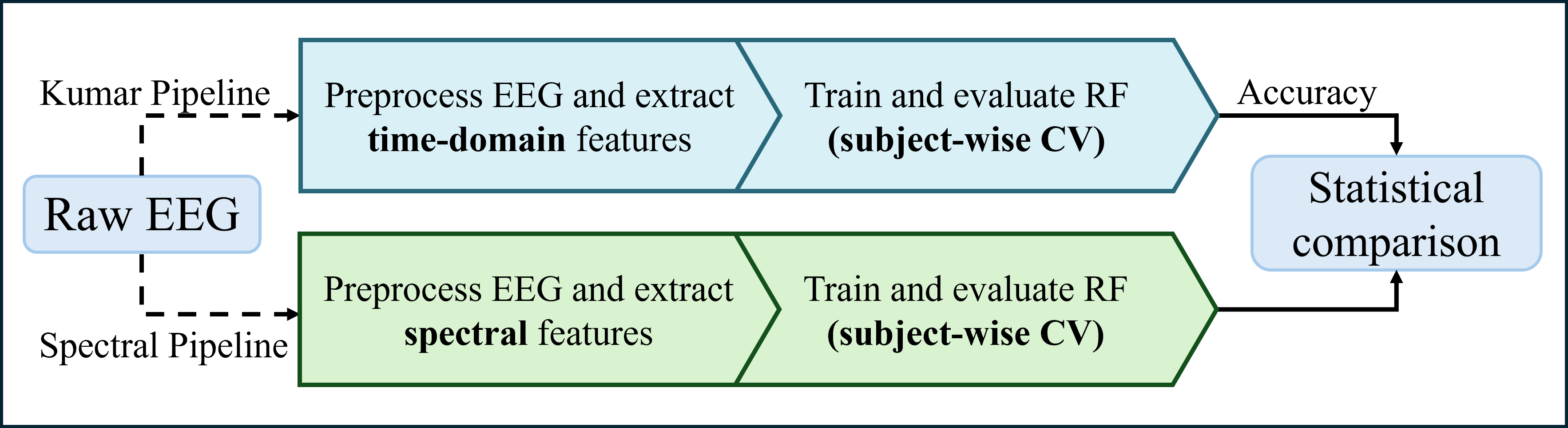}
    \caption{Conceptual overview of the pipeline comparison approach. The diagram goes from left to right. The dataset is preprocessed and feature-extracted in two different manner, but cross-validation (CV) is identical. Finally, their performances are compared statistically.}
    \label{fig:method_overview}
\end{figure*}

Accordingly, this work aims to establish a transparent and reproducible framework for imagined speech decoding on Kumar’s EEG dataset under a strictly subject-independent evaluation protocol. Two distinct preprocessing and feature-extraction pipelines are compared with respect to their downstream classification performance using an RF classifier. The first pipeline is based on the statistical time-domain feature framework proposed by \cite{Kumar}, while the second pipeline focuses on frequency-domain spectral features. Both pipelines are evaluated using subject-wise cross-validation (CV) and trial-level majority voting. To the best of our knowledge, this study provides the first evaluation of the Kumar dataset under these constraints, offering a realistic reference point for future methodological advances with respect to cross-subject generalization.

\section{Methodology}

This section outlines the key methods employed in the study and includes an explanation of the comparison framework, details of Kumar’s EEG dataset, and finally a description of each pipeline.   

\subsection{Proposed Method}

This study compares two different preprocessing and feature extraction pipelines, denoted as the Kumar Pipeline and the Spectral Pipeline. A general overview of the pipeline comparison is illustrated in Fig. \ref{fig:method_overview} to clarify the methodological approach.

\subsection{Dataset}

\begin{figure}[b]
    \centering
    \includegraphics[width=0.85\linewidth]{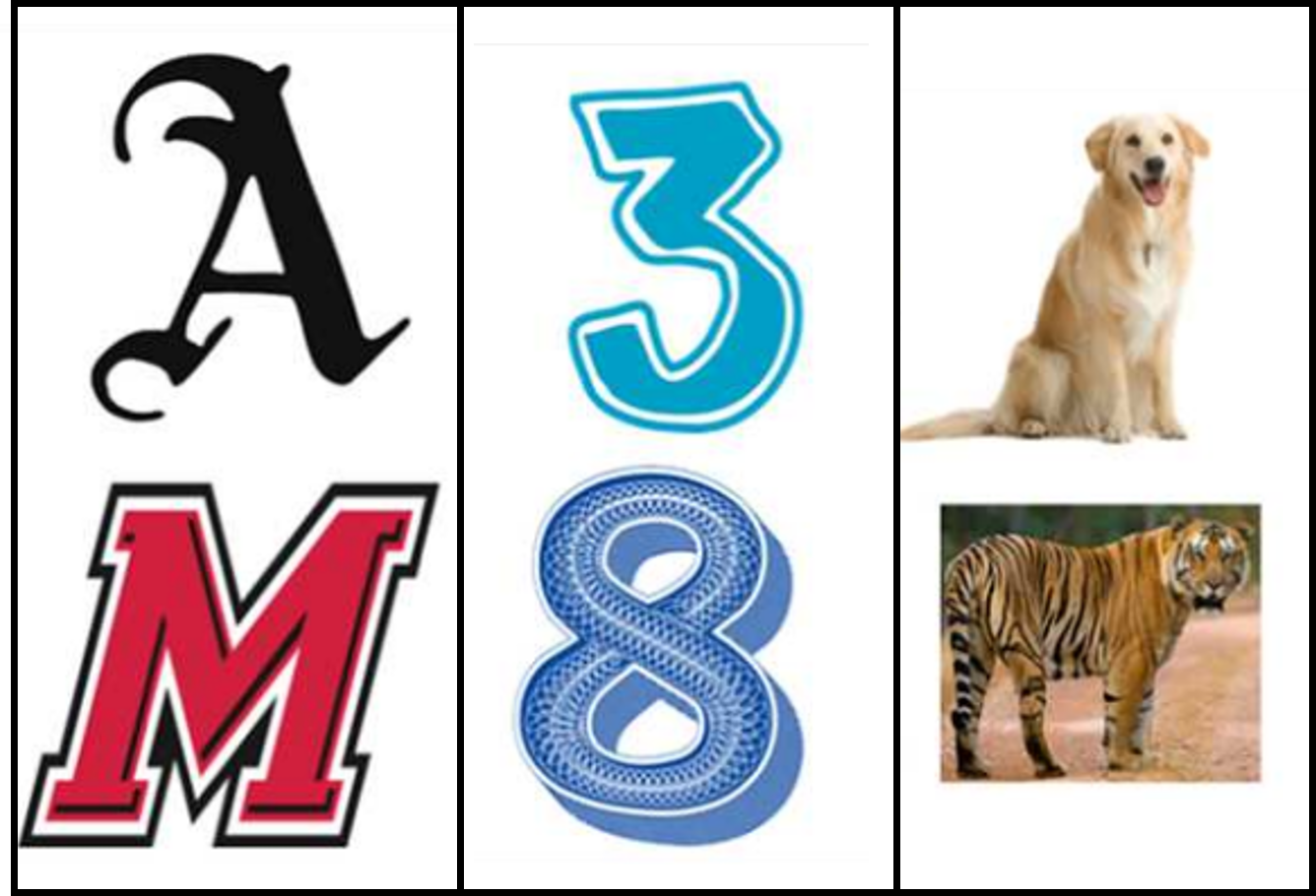}
    \caption{Examples of stimulus within the coarse-grained classes—characters, digits, and images—presented from left to right. These images are adapted from \cite{Kumar}.}
    \label{fig:Classes}
\end{figure}

The data, as described in \cite{Kumar}, were collected using Emotiv EPOC+ with 14 channels (AF3, F7, F3, FC5, T7, P7, O1, O2, P8, T8, FC6, F4, F8, and AF4) placed according to the international 10-20 system, with DRL and CMS, positioned above the ears, serving as reference electrodes. The EEG was captured at 2048 Hz and later down-sampled to 128 Hz. Only the down-sampled data is publicly available. The experimental setup was as follows: Each participant looked at a computer screen where an object was presented, followed by 10 s of imagination with closed eyes. A 20-second rest period was provided between imagination trials to allow the participant to return to a resting state. EEG data from 23 subjects aged 15-40 were collected while imagining the following targets: characters (A,C,F,H,J,M,P,S,T,Y), digits (0-9) and images from everyday life (apple, car, dog, ring, phone, rose,  scooter, tiger, wallet, and watch). Fig \ref{fig:Classes} illustrates representative examples from each class. A total of 690 ($23\times30$) EEG recordings were obtained, each lasting 10 seconds. Note that this work mainly focuses on coarse-level classification, discriminating between characters, digits, and images.

\subsection{Common Preprocessing and Segmentation}

This section describes the preprocessing and segmentation steps shared across both pipelines. Inspection of the publicly available dataset revealed several inconsistencies relevant to data handling, which are briefly summarized here. Each trial contains 12 s of EEG data without explicit markers indicating trial onset; therefore, the central 10 s segment was retained for analysis. Additionally, folders of the digit condition contain indices ranging from 0–24, with indices 2 and 18 missing, whereas the character and image conditions span from 0–22. As the dataset documentation consistently reports 23 subjects, these discrepancies were attributed to indexing errors. Accordingly, all folders were manually re-indexed prior to data loading, without removing or modifying any trials.

Following pipeline-specific filtering, which will be described in the subsequent sections, each 10 s trial was segmented into 1 s windows with 75\% overlap, with each window containing data from all 14 channels. A window length of 1 s provides an appropriate trade-off between temporal resolution and frequency resolution for subsequent spectral feature extraction. Prior to classification, subject-wise z-score normalization was applied in accordance with the subject-independent CV scheme and is defined as:

\begin{align}
z &= \frac{x - \mu}{\sigma},
\label{Eq:zscore}
\end{align}

where $x$ denotes the feature value, $z$ the normalized feature, and $\mu$ and $\sigma$ are the subject-wise mean and standard deviation of the given feature, respectively.

\subsection{Kumar Pipeline}

This pipeline closely follows the feature extraction approach proposed by Kumar et al.\cite{Kumar}. The down-sampled EEG data (128 Hz) were first smoothed using a 5-point moving average filter. Trials were then segmented as described previously; the use of 1 s overlapping windows differs from the original pipeline and was introduced to ensure comparability with the spectral pipeline.

Four statistical features—sd, rms, sum, and signal energy—were computed per channel for each window, resulting in a 56-dimensional feature vector (14 channels × 4 features) and a feature matrix of size $27600 \times 56$.

Inspection of the extracted features revealed that the sd exhibited a positively skewed distribution; therefore, a logarithmic transformation was applied prior to normalization. Finally, all features were normalized as described in Section II-C.

\subsection{Spectral Pipeline}

\begin{figure}
    \centering
    \fbox{\includegraphics[width=0.95\linewidth]{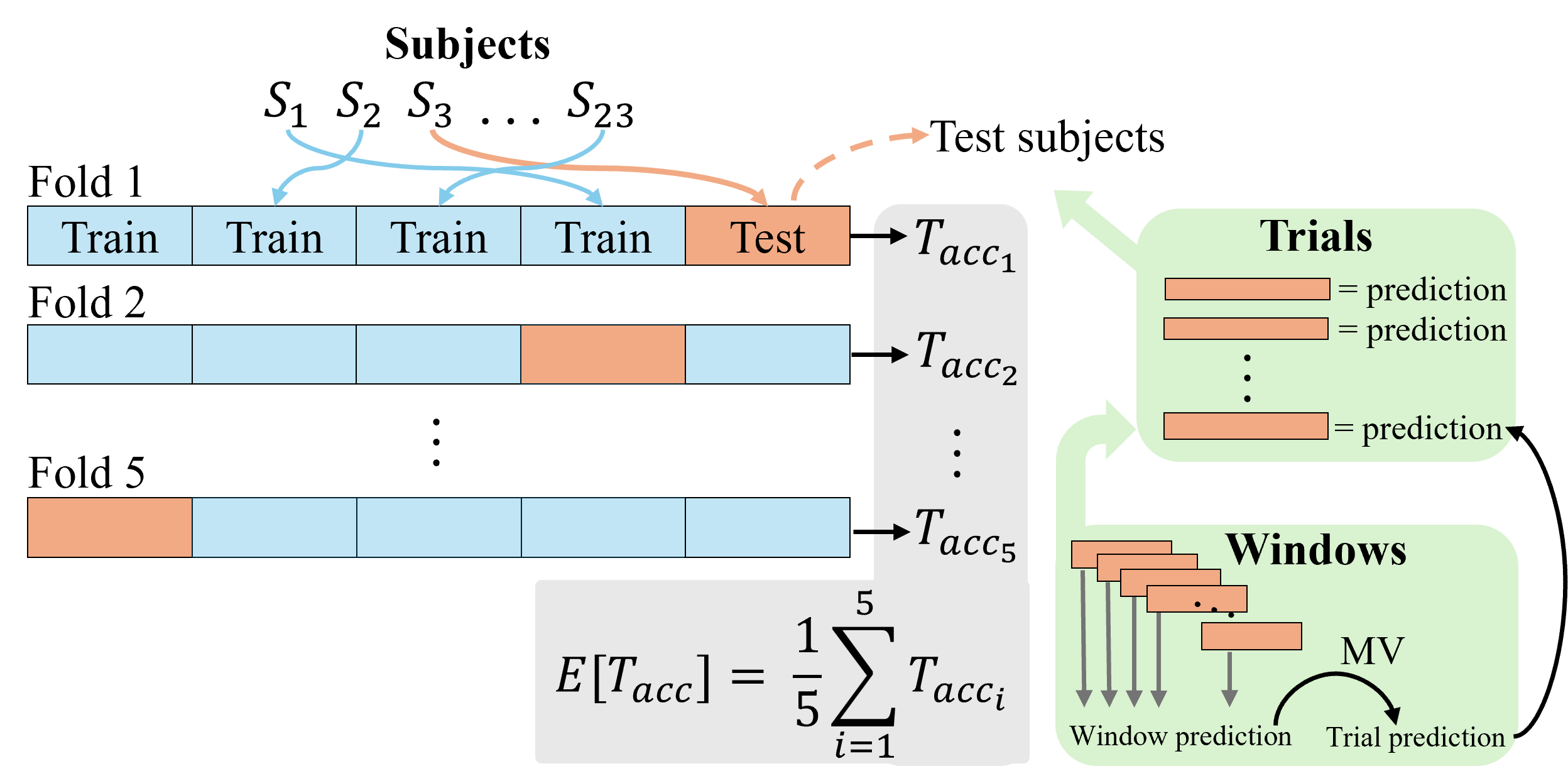}}
    \caption{Schematic overview of the subject-wise cross-validation and trial-level evaluation procedure. In each fold, models are trained on a subset of subjects and tested on held-out subjects. Window-level predictions are aggregated to the trial level using majority voting (MV), and final performance is obtained by averaging trial-wise accuracies across folds.}
    \label{fig:5foldCV}
\end{figure}

The raw EEG data were first filtered with a 50 Hz finite impulse response notch filter with zero phase to remove powerline interference, while preserving phase characteristics. A zero-phase 4th-order infinite impulse response Butterworth bandpass filter (0.5–60 Hz) was applied to reduce slow drift while preserving the temporal structure of the EEG signal, which is required for accurate fast-Fourier transform-based spectral feature estimation. Following filtering, trials were segmented as described previously.

Spectral features were extracted by estimating the power spectral density (PSD) of each window using a periodogram approach. Absolute band power was computed by numerically integrating the PSD within each frequency band using the trapezoidal rule:

\begin{align}
P_{b} &= \sum_{f \in b} S_{xx}(f)\,\Delta f, \label{Eq:Power}
\end{align}

where $P_{b}$ denotes the power within frequency band $b$, $S_{xx}(f)$ is the PSD estimate at frequency $f$, and $\Delta f$ is the frequency resolution. Frequency bands were defined as half-open intervals to avoid overlap: delta [0.5, 4) Hz, theta [4, 8) Hz, alpha [8, 14) Hz, beta [14, 30) Hz, and gamma [30, 64] Hz.\cite{bands1,bands2}. The upper limit of 64 Hz for the gamma band is determined by the Nyquist theorem, given the sampling frequency of 128 Hz \cite{Sada}. As band-power features typically exhibit positively skewed distributions, a logarithmic transformation was applied to reduce skewness, followed by z-score normalization. Each window was represented by a 70-dimensional feature vector (14 channels × 5 features), resulting in a feature matrix of size $27600 \times 70$ after aggregation across all trials.

\subsection{Evaluation and Statistical Analysis}

Classification was performed using an RF classifier, consistent with the original work of Kumar et al.\cite{Kumar}. The RF employed bootstrap aggregation, majority voting across decision trees, and the Gini index as the impurity criterion. The number of decision trees was fixed at 200, while all remaining hyperparameters were set to their default values in scikit-learn.

Model performance was evaluated using a 5-fold CV scheme with subject-wise splits to ensure subject independence (see Fig. \ref{fig:5foldCV}). Stratification and grouping by subject were implemented using Python scikit-learns \texttt{StratifiedGroupKFold}, which ensures constant class distribution for the training and test folds. Within each test fold, window-level predictions were aggregated to the trial level using majority voting. Accuracy for each test fold was defined as the trial-level accuracy across all test subjects. Final model performance was obtained by averaging trial-wise accuracies across the five folds. For model comparison, trial-wise accuracies were first averaged within each subject to avoid subject-wise correlation between predictions. The resulting mean subject-wise accuracies were compared between pipelines using a paired Wilcoxon signed-rank test, which is recommended for classifier comparisons \cite{Demsar2006}. Let \( d_i = a_i^{(1)} - a_i^{(2)} \) denote the difference in mean subject-wise trial accuracies between the two pipelines for subject \( i \), i.e. $a_i^{(1)}$ and $a_i^{(2)}$. The Wilcoxon signed-rank test statistic is defined as:
\begin{align}
T &= \sum_{i=1}^{N} \operatorname{sgn}(d_i)\,R_i,
\label{Eq:wilcoxon}
\end{align}
where \( R_i \) is the rank of \( |d_i| \) among all non-zero absolute differences \( \{|d_1|,\dots,|d_N|\} \), \( N \) is the number of subjects, and \( \operatorname{sgn}(\cdot{})
 \) denotes the sign function \cite{Wilcox1}.

\subsection{Forward Feature Selection}

In addition to the pipeline performance comparison, a forward feature selection procedure was applied to the spectral pipeline to assess the contribution of individual frequency bands to classification performance \cite{Brockhoff}. Starting from single-band feature sets, bands were incrementally added based on improvements in trial-wise classification accuracy, using the same subject-wise CV and evaluation protocol as described above.

\section{Results}

While the Kumar's EEG dataset also supports fine-level classification, the present study focuses on coarse-level decoding. The mean trial-wise accuracies for coarse-level prediction are reported in Table~\ref{tab:final_results}. The Spectral Pipeline achieved a higher accuracy than the Kumar Pipeline (49.03 ± 4.18 \% vs. 37.97 ± 3.79 \%). The difference in pipeline performance was statistically significant according to the paired Wilcoxon signed-rank test (p = 0.018), with a large effect size (rank-biserial correlation = 0.57). Table \ref{tab:classwise_accuracy} displays the class-wise accuracies for each pipeline. Both pipelines achieved the highest accuracy for the character class, while performance for digits and images was lower.

\begin{table}[htbp]
\centering
\caption{Final evaluation of pipeline performance}
\label{tab:final_results}
\begin{tabular}{lc}
\hline
Pipeline & Accuracy (\%) \\
\hline
Spectral Pipeline & $\mathbf{49.03 \pm 4.18}$ \\
Kumar Pipeline    & $37.97 \pm 3.79$ \\
\hline
\end{tabular}
\end{table}

\begin{table}[htbp]
\centering
\caption{Class-wise classification accuracy for each pipeline.}
\label{tab:classwise_accuracy}
\begin{tabular}{lcc}
\hline
Class & Kumar Pipeline & Spectral Pipeline \\
\hline
Character & $\mathbf{48.90 \pm 9.19}$ & $\mathbf{56.30 \pm 8.94}$ \\
Digit & $36.90 \pm 8.70$ & $46.00 \pm 10.24$ \\
Image & $29.40 \pm 14.16$ & $48.50 \pm 13.94$ \\
\hline
\end{tabular}
\end{table}

A forward selection procedure was applied to determine which individual frequency band, or combination of bands, yielded the highest classification accuracy. As depicted by Table \ref{tab:forward_selection_bands}, the alpha band achieved the highest single-band mean accuracy. Adding the delta band resulted in a substantial performance increase, while the highest mean accuracy was obtained using four bands ($\alpha$+$\delta$+$\gamma$+$\theta$), excluding beta. All band combinations were statistically compared against the full-band model using the Wilcoxon test. A Bonferroni correction was applied to account for multiple comparisons, and no statistically significant differences were observed.

\begin{table}[ht]
\caption{Stepwise forward selection of frequency bands. Bold values indicate mean accuracies exceeding the full-band model.}
\label{tab:forward_selection_bands}
\centering
\begin{tabular}{c l c}
\hline
Step & Band set & Accuracy (\%) \\
\hline
1 & $\alpha$ & $46.67 \pm 6.93$ \\
2 & $\alpha + \delta$ & $\mathbf{51.17 \pm 5.66}$ \\
3 & $\alpha + \delta + \gamma$ & $\mathbf{51.03 \pm 6.58}$ \\
4 & $\alpha + \delta + \gamma + \theta$ & $\mathbf{52.83 \pm 4.64}$ \\
5 & $\alpha + \delta + \gamma + \theta + \beta$ & $49.03 \pm 4.18$ \\
\hline
\end{tabular}
\end{table}

\section{Discussion}

The Spectral Pipeline outperformed the Kumar Pipeline, indicating that frequency-domain features generalize better across subjects than time-domain statistical features in this setting. Forward feature selection showed that this improvement could not be explained by the higher feature dimensionality, as a comparable performance was achieved using only a subset of frequency bands (see Table \ref{tab:forward_selection_bands}). Both pipelines achieved higher accuracy for characters than for digits and images, indicating class-dependent differences in signal discriminability (see Table \ref{tab:classwise_accuracy}). Absolute classification accuracies were lower than those reported by Kumar et al. \cite{Kumar}. This was expected given the use of a strictly subject-independent evaluation protocol, which provides a more realistic estimate of generalization performance for practical BCI implementation. Without careful and realistic evaluation, models that seemingly perform well may still fail in real-world scenarios. The forward selection analysis further indicated that not all frequency bands contribute equally to class separation. Alpha-band power showed the highest discriminative value, potentially reflecting the closed-eyes imagination task. The second-highest mean trial-wise accuracy in Table \ref{tab:forward_selection_bands} was found using only the alpha and delta band. This was not significantly different from the full-band model, suggesting that much of the relevant information for coarse-level imagined speech classification can be captured by a limited set of frequency bands.

The minimal artifact removal in the pipelines is a limitation of this work, as the removal of artifacts improves data quality and enables a better representation of relevant brain activity as a foundation for ML \cite{Alzahrani2024Review}. Channel selection was not performed in this work and may prove to be of interest, as prior imagined speech studies have shown that selecting a subset of task-relevant EEG channels can enhance discriminative performance \cite{Panachakel2020}. Cortical regions may contribute differently depending on the given task, and within this experimental framework, visual imagery might engage the occipital electrodes to a higher degree. Since the Spectral Pipeline only used absolute band power as spectral features, future work may benefit from including additional spectral features, as in \cite{Chengaiyan2020}, which may provide further discriminative information for imagined speech decoding.

\section{Conclusion}

This study established a transparent subject‑independent investigation of imagined speech decoding using the widely adopted Kumar EEG dataset. In this work, a time-domain statistical feature pipeline was compared with a frequency-domain band-power feature pipeline under a strict subject-wise CV. This provided a realistic assessment of cross-subject generalization, which is essential for practical BCI applications. The spectral pipeline consistently outperformed the statistical approach, demonstrating that frequency‑domain representations offer more robust and transferable information for coarse‑level imagined speech classification.

Forward feature selection further revealed that a limited subset of frequency bands, particularly alpha and delta captured most of the discriminative structure in the data, indicating that improved generalization does not require high‑dimensional feature spaces. The observed class‑dependent performance differences also highlight the inherent variability in the discriminability of imagined speech categories.

Overall, this work offers a reproducible and clearly defined baseline for future research on subject‑independent imagined speech decoding. The findings underscore the importance of feature representation when evaluation protocols are aligned with real‑world constraints and provide a foundation for developing more advanced, generalizable EEG‑based BCI systems.

\end{document}